\documentclass[twocolumn, switch]{article} 

\usepackage{preprint}
\usepackage[dvipsnames, table]{xcolor}
\usepackage{amsmath, amsthm, amssymb, amsfonts}
\usepackage{bm}
\usepackage{amsmath}
\usepackage{graphicx}
\usepackage{enumerate}
\usepackage[algoruled,algo2e]{algorithm2e}

\usepackage{setspace}
\usepackage{amssymb}
\usepackage{booktabs}
\usepackage{array}
\usepackage{tabularx}
\newcolumntype{Y}{>{\centering\arraybackslash}X}
\usepackage{wrapfig}
\usepackage{color}
\usepackage{multirow}
\usepackage{multicol}
\usepackage{threeparttable}
\usepackage{url}
\usepackage[page]{appendix}

\usepackage[numbers,square]{natbib}

\usepackage[utf8]{inputenc}	
\usepackage[T1]{fontenc}	
\usepackage[colorlinks = true,
            linkcolor = blue,
            urlcolor  = blue,
            citecolor = blue,
            anchorcolor = black]{hyperref}	
\usepackage{booktabs} 		
\usepackage{nicefrac}		
\usepackage{microtype}		
\usepackage{lineno}		
\usepackage{float}			

\usepackage{newfloat}
\DeclareFloatingEnvironment[name={Supplementary Figure}]{suppfigure}
\usepackage{sidecap}
\sidecaptionvpos{figure}{c}
\usepackage{graphicx}
\usepackage{subcaption}
\usepackage{amssymb}

\usepackage{algorithm}
\usepackage{algorithmic}
\definecolor{Gray}{gray}{0.93}

\usepackage{titlesec}
\titlespacing\section{0pt}{12pt plus 3pt minus 3pt}{1pt plus 1pt minus 1pt}
\titlespacing\subsection{0pt}{10pt plus 3pt minus 3pt}{1pt plus 1pt minus 1pt}
\titlespacing\subsubsection{0pt}{8pt plus 3pt minus 3pt}{1pt plus 1pt minus 1pt}

\usepackage{graphics}
\usepackage{hyperref}
\usepackage{color}
\usepackage{multirow}
\usepackage{hhline}
\usepackage{lipsum}
\usepackage{pifont}
\usepackage{amsfonts} 
\usepackage{colortbl} 

\usepackage{amsmath}

\usepackage{multirow}

\newtheorem{theorem}{Theorem}

\newtheorem{proposition}[theorem]{Proposition}

\title{ETC: Extreme Token Compression via Task-aware Visual Information Distillation in VLMs}
\usepackage{authblk}

\author[1]{Yiling Gao}
\author[1]{Hongchen Wei}
\author[*1]{Zhenzhong Chen}
\affil[1]{School of Remote Sensing and Information Engineering, Wuhan University}

\begin{document}

\twocolumn[ 
  \begin{@twocolumnfalse} 
  
\maketitle

\begin{abstract}

In Vision-Language Models (VLMs), high-resolution images produce a large number of visual tokens, resulting in high computational costs and KV-cache overhead during inference. To address this problem, we propose an \textbf{Extreme Token Compression (ETC)} framework that minimizes task loss when reducing the number of input tokens based on the principle of variational information distillation. Specifically, from an information-theoretic perspective, we show that minimizing task loss requires the compact representation to preserve the instruction-aware sufficient statistic of the task-relevant visual information for prediction. In practice, ETC leverages text-to-image cross-attention to weight the original visual features to approximate the latent instruction-aware predictive statistic. Moreover, ETC introduces a variational information distillation, enabling the compact representation to preserve the essential information to recover this predictive statistic. Experiments on LLaVA-1.5-7B and Qwen3-VL-2B show that ETC remains effective even under single-token compression, substantially reducing KV-cache overhead while retaining strong task performance.

\end{abstract}

\vspace{0.4cm}

  \end{@twocolumnfalse} 
] 

\newcommand\blfootnote[1]{%
\begingroup
\renewcommand\thefootnote{}\footnote{#1}%
\addtocounter{footnote}{-1}%
\endgroup
}

\section{INTRODUCTION}
\label{sec:introduction}
{\blfootnote{\scalebox{0.98}{Corresponding author: Zhenzhong Chen, E-mail:zzchen@ieee.org}}}
Recent VLMs \cite{yang2025qwen3,wang2025internvl3,xu2025llava} consist of a visual encoder \cite{li2025tokenpacker}, a projector, and a pretrained LLM. These models have significantly improved visual reasoning and high-resolution image understanding. However, these capabilities also lead to a larger number of visual tokens. For example, a standard $336 \times 336$ image already yields $576$ visual tokens in LLaVA-1.5 \cite{liu2024improved}, whereas high-resolution models like Qwen3-VL \cite{yang2025qwen3} can generate over $10{,}000$ visual tokens. Since Transformer computation scales quadratically with sequence length \cite{vaswani2017attention}, these visual tokens result in higher computational and KV-cache costs, making visual token compression an important research topic.

Existing methods mainly reduce visual tokens in two ways: selective compression and abstractive compression. Selective methods, such as SparseVLM \cite{zhang2025sparsevlm} and TopV \cite{yang2025topv}, discard or merge image patches using heuristics like attention scores or spatial similarity. They can work well under moderate compression, but often suffer from a ``compression cliff'' \cite{shang2025llava,ye2025atp}, where performance drops sharply once the token budget becomes very small. This suggests that simply retaining the seemingly most important patches is unreliable in the extreme compression regime.

On the other hand, abstractive methods summarize the full visual sequence into a few compact tokens. Representative examples include MQT-LLAVA \cite{hu2024matryoshka} and VoCo-LLaMA \cite{ye2025voco}. Compared with selective methods, they are better suited to synthesizing information across the image. However, most of them are trained only through Next-Token Prediction (NTP) loss \cite{yang2025pvc}, which provides only indirect supervision for compression. As long as the final text prediction is correct, the model is not explicitly required to preserve the visual information needed for the task, and may rely too heavily on language priors.

In this work, we study visual token compression from the perspective of task-loss minimization, with the goal of retaining the most useful visual information for downstream prediction. Based on this insight, we propose an
\textbf{Extreme Token Compression (ETC)} framework, which distills the task-relevant visual information into the compact representation to satisfy the task-loss minimization requirement. In theory, ETC derives a guideline that the compact representation should preserve the instruction-aware sufficient statistic of the original visual tokens if the task loss is minimized under compression. In practice, it utilizes text-to-visual cross-attention to weight the original visual tokens to approximate the latent instruction-aware predictive statistic, and introduces a variational information distillation objective~\cite{ahn2019variational} to preserve this approximation in the compact representation. Implemented in a decoder-only VLM with a bottleneck compression attention mask similar to VoCo~\cite{ye2025voco}, ETC guides the compact representation to preserve the crucial task-relevant visual information directly, maintaining substantial effectiveness even under extreme compression (e.g., 1--4 tokens).

The rest of this paper is organized as follows. We first introduce the inference costs and prior work on visual token compression in Section~\ref{sec:related_work}. In Section~\ref{sec:method}, we derive the theoretical requirements for task-loss minimization and present their practical implementation for token compression. Finally, the proposed method is validated and discussed in Sections~\ref{sec:experiments} and~\ref{sec:discussion}.

\section{Related Work}
\label{sec:related_work}
The sequence length of visual tokens is a key factor in the efficiency of VLMs. This is because the computational cost of the self-attention mechanism scales quadratically as $O(N^2)$ \cite{belhaouari2025efficient,zeng2025dylofvit} with the number of tokens. To reduce the number of tokens, InternVL 3.5 \cite{wang2025internvl3} generates 256 tokens per $448 \times 448$ tile after a $2\times2$ Pixel Unshuffle operation. However, to preserve fine-grained detail, modern VLMs generate increasingly more image tokens to achieve higher performance. For example, LLaVA-v1.5 \cite{li2025llavaonevision} produces 576 tokens per image. Qwen3-VL \cite{ahmed2025qwen} utilize dynamic resolution that can result in over $10,000$ tokens. This growth in sequence length causes computational cost and KV-cache storage to increase \cite{behnam2025rocketkv,wu-etal-2025-scope}. However, high-resolution inputs often contain redundancy that contributes less to downstream reasoning \cite{zhao2025accelerating}. Therefore, visual token compression is a promising direction for reducing the cost of VLM inference while retaining strong performance.

Existing visual token compression work mainly falls into two paradigms. Selective compression preserves semantically significant visual tokens while reducing redundancy. Pruning-based methods such as ST3 \cite{zhuang2025st3}, LVPruning \cite{sun2025lvpruning}, and Fit-and-Prune \cite{ye2025fit} discard background tokens using importance-driven metrics, while merging-based approaches such as PACT \cite{dhouib2025pact}, TempMe \cite{shen2025tempme}, and LTM \cite{wang2025efficient} consolidate visually similar patches through grouping or clustering. Under aggressive compression, these methods can incur irreversible information loss or feature over-smoothing. Consequently, such techniques are typically limited to a 50\%-70\% compression rate to maintain a reasonable balance between efficiency and accuracy. Abstractive compression instead condenses large-scale visual sequences into a few latent representations. Query-based resampling methods, such as HierarQ \cite{11093233}, PQR \cite{amoroso2025perceive}, and LLaMA-Vid \cite{li2024llama}, pool features into a compact set of tokens via cross-attention, while VoCo-LLaMA \cite{ye2025voco} performs internal distillation through bottleneck attention masks. However, these methods are typically trained only through next-token prediction loss, leaving the preservation of task-relevant visual information weakly supervised and allowing the model to rely on textual priors rather than faithfully retaining visual evidence.

\section{Methodology}
\label{sec:method}

\begin{figure*}[t]
  \vskip 0.2in
  \centering
    \centerline{\includegraphics[width=\textwidth]{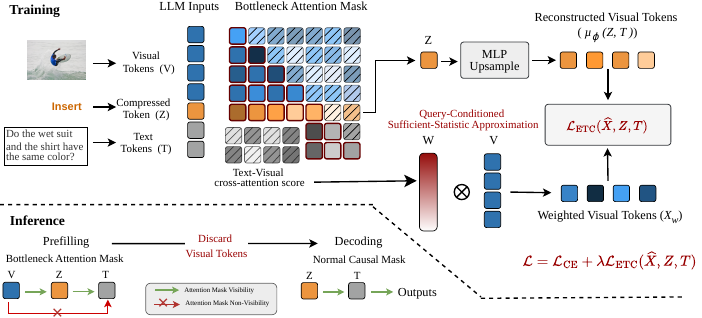}}
    \caption{
        Overview of ETC. \textbf{Training:} compressed tokens $Z$ are inserted between visual tokens $V$ and text tokens $T$, and a bottleneck attention mask allows $Z$ to aggregate information from $V$ while preventing $T$ from directly attending to the raw visual tokens. At the final LLM layer, text-to-image cross-attention scores produce instruction-aware weights that define the predictive-statistic estimate $\widehat{X}$; in parallel, an MLP decoder maps $Z$ to $\mu(Z,T)$, and the variational information distillation loss encourages the reconstructed compressed representation to preserve this estimate. \textbf{Inference:} after prefilling with the same mask, $V$ is discarded from the KV cache and decoding uses only $Z$ and $T$ under a normal causal mask.
    }
    \label{method}
\end{figure*}

\subsection{Task-Loss Minimization}

Let $V \in \mathcal{V}$ be the original visual token sequence, $T \in \mathcal{T}$ the textual query or instruction, and $Y \in \mathcal{Y}$ the target output. ETC compresses $V$ into a compact instruction-aware representation:
\begin{equation}
Z = f_\theta(V,T),
\label{eq:compressor}
\end{equation}
where $Z \in \mathcal{Z}$ is the compact representation and $f_\theta$ is a compressor, which compresses the visual token sequence $V$ into a compact representation $Z$ guided by the instruction $T$. The compressed representation $Z$ is then used in place of $V$ for downstream prediction.

We define the compression loss as the excess task loss incurred by using $Z$ instead of $V$ for prediction. Under the log-likelihood objective, this loss is formulated as:
\begin{equation}
\Delta_{\mathrm{task}}(Z)
=
H(Y\mid Z,T)-H(Y\mid V,T).
\label{eq:task_gap}
\end{equation}
where $H(\cdot\mid\cdot)$ denotes conditional entropy and $I(\cdot;\cdot\mid\cdot)$ denotes conditional mutual information.

\begin{proposition}[Criterion for task-loss minimization]
\label{prop:task_lossless_sufficiency}
According to Eq.~\eqref{eq:task_gap}, the task loss caused by the compressor $f_\theta$ can be formulated as:
\begin{equation}
\Delta_{\mathrm{task}}(Z)
=
I(Y;V\mid Z,T).
\label{eq:task_lossless_mi}
\end{equation}
where $I(\cdot;\cdot\mid\cdot)$ denotes conditional mutual information.
Moreover,
\begin{equation}
\Delta_{\mathrm{task}}(Z)=0
\iff
Y \perp V \mid (Z,T),
\label{eq:task_lossless_ci}
\end{equation}
which implies that minimizing the compression loss requires $Z$ to be a sufficient statistic of $V$ for predicting $Y$ conditioned on the instruction $T$.
\end{proposition}

Accordingly, Eq.~\eqref{eq:task_lossless_ci} characterizes an instruction-aware predictive statistic:
\begin{equation}
X^\star = g^\star(V,T) := p(\cdot \mid V,T).
\label{eq:ideal_statistic}
\end{equation}
where $g^\star$ denotes the mapping from $(V,T)$ to the predictive sufficient statistic, and $p(\cdot \mid V,T)$ denotes the conditional distribution of $Y$ given $V$ and $T$.
According to Eq.~\eqref{eq:task_lossless_ci}, ideal compression satisfies $p(\cdot\mid V,T)=p(\cdot\mid Z,T)$. So we have
\begin{equation}
H(X^\star\mid Z,T)=0.
\label{eq:ideal_psc}
\end{equation}
Eq.~\eqref{eq:ideal_psc} states that the compact representation $Z$, together with the query $T$, retains all information needed to determine the latent predictive statistic $X^\star$. Thus, the sufficient statistic $Z$ is realized by preserving a latent instruction-aware predictive statistic $X^\star$. Detailed derivations are deferred to Appendix~\ref{app:recoverable_statistic}.

In practice, however, $X^\star$ represents an output distribution and cannot be directly accessed. We therefore address the task-loss minimization via a proxy statistic $\widehat{X}=g(V,T)\approx X^\star$, which approximates the latent predictive statistic and should be preserved in $Z$. Therefore, we have
\begin{equation}
H(\widehat{X}\mid Z,T)\rightarrow 0.
\label{eq:psc_practical}
\end{equation}

\subsection{Cross-Attention for Instruction-aware Predictive-Statistic Approximation}

The proxy statistic $\widehat{X}=g(V,T)\approx X^\star$ needs an instruction-aware mapping within the model architecture. In a decoder-only VLM, visual information affects the textual computation through text-to-image cross-attention. Let $X_v=[x_1,\ldots,x_{N_v}]\in\mathbb{R}^{N_v\times D}$ be the visual features at the selected LLM layer, where $N_v$ is the number of visual tokens and $D$ is the hidden dimension. For the $j^{th}$ text token, the visual contribution can be written as
\begin{equation}
o_j=\sum_{i=1}^{N_v} A_{j,i} W_V x_i,
\label{eq:cross_attn_output}
\end{equation}
where $A_{j,i}$ is the attention score from the $j^{th}$ text token to $i^{th}$ visual token, $W_V$ is the value-weight matrix. This makes cross-attention a natural mapping, since it determines how visual tokens contribute to the text-side representations used to predict \(Y\).

Let $A^h_{j,i}$ denote the attention from $j^{th}$ text token to $i^{th}$ visual token at head $h$. ETC weights the $i^{th}$ visual token by the mean text-to-image cross-attention score across all heads and text tokens:
\begin{equation}
S_i = \frac{1}{N_h\cdot N_t}\sum_{h=1}^{N_h}\sum_{j=1}^{N_t} A^h_{j,i},
\label{eq:tg_score_raw}
\end{equation}
where $N_h$ is the number of attention heads and $N_t$ is the number of text tokens. We then apply min-max normalization over cross-attention scores $S=\{S_i\}_{i=1}^{N_v}$:
\begin{equation}
\tilde{S}_i =
\begin{cases}
\dfrac{S_i-\min(S)}{\max(S)-\min(S)}, & \text{if } \max(S)>\min(S),\\[0.8em]
1, & \text{otherwise}.
\end{cases}
\label{eq:tg_score_norm}
\end{equation}

The instruction-aware predictive-statistic estimate is represented as
\begin{equation}
\widehat{X} = (1-\alpha)X_v + \alpha(X_v \odot \tilde{S})
= X_v \odot (1-\alpha+\alpha\tilde{S}),
\label{eq:xw_def}
\end{equation}
where $\tilde{S}=\{\tilde{S}_i\}_{i=1}^{N_v}$ is broadcast along the feature dimension, $\odot$ denotes element-wise multiplication, and $\alpha\in[0,1]$ controls the contribution of instruction-aware weighting.

\subsection{Variational Information Distillation for Predictive-Statistic Preservation}

Eq.~\eqref{eq:psc_practical} requires the compact representation to preserve the proxy instruction-aware predictive statistic, which amounts to minimizing the conditional entropy $H(\widehat{X}\mid Z,T)$. Using the conditional mutual information identity,
\begin{equation}
I(\widehat{X};Z\mid T)=H(\widehat{X}\mid T)-H(\widehat{X}\mid Z,T),
\label{eq:mi_xw_z}
\end{equation}
$H(\widehat{X}\mid T)$ does not depend on Z, it can be treated as a constant during optimization. So reducing $H(\widehat{X}\mid Z,T)$ is equivalent to maximizing the mutual information between $\widehat{X}$ and $Z$ given the query. Exact optimization of $I(\widehat{X};Z\mid T)$ is not directly available because the true conditional distribution $p(\widehat{X}\mid Z,T)$ is unknown. Following Variational Information Distillation (VID)~\cite{ahn2019variational}, we introduce a variational distribution $q(\widehat{X}\mid Z,T)$ to approximate $p(\widehat{X}\mid Z,T)$, which yields the lower bound
\begin{equation}
I(\widehat{X};Z\mid T)
\ge
H(\widehat{X}\mid T)+\mathbb{E}_{p(\widehat{X},Z,T)}\big[\log q(\widehat{X}\mid Z,T)\big].
\label{eq:variational_lower_bound}
\end{equation}
where $\mathbb{E}_{p(\widehat{X},Z,T)}[\cdot]$ denotes expectation under the joint distribution of $\widehat{X}$, $Z$, and $T$. Since $H(\widehat{X}\mid T)$ does not depend on the model parameters, maximizing this bound is equivalent to minimizing
\begin{equation}
\mathcal{L}_{\mathrm{ETC}}
=
-\mathbb{E}_{p(\widehat{X},Z,T)}\big[\log q(\widehat{X}\mid Z,T)\big],
\label{eq:suf_general}
\end{equation}
which follows the same variational objective as VID.

Following the Gaussian parameterization used in VID, we model the variational distribution with heteroscedastic mean $\mu(Z,T)$ and homoscedastic variance shared across samples:
\begin{equation}
q(\widehat{X}\mid Z,T)
=
\prod_{n=1}^{N}\prod_{d=1}^{D}
\mathcal{N}\!\big(\widehat{X}_{n,d};\mu_{n,d}(Z,T),\sigma_d^2\big),
\label{eq:gaussian_decoder}
\end{equation}
where $\mu_{n,d}(Z,T)$ denotes the $(n,d)$-th element of the predicted mean, and $\sigma_d^2$ is the learnable variance. In practice $N=N_v$, and $\mu(Z,T)$ is aligned with $\widehat{X}\in\mathbb{R}^{N\times D}$. Substituting Eq.~\eqref{eq:gaussian_decoder} into Eq.~\eqref{eq:suf_general} and dropping constants gives the objective
\begin{equation}
\mathcal{L}_{\mathrm{ETC}}
=
\frac{1}{B}\sum_{b=1}^{B}\left[
\sum_{d=1}^{D}\left(
\frac{1}{2\sigma_d^2}\sum_{n=1}^{N}
\big(\widehat{X}_{b,n,d}-\mu_{b,n,d}(Z,T)\big)^2
+ N\log \sigma_d
\right)
\right],
\label{eq:suf_gaussian}
\end{equation}
where where $B$ is the size of batch and $b\in\{1,\ldots,B\}$ indexes training samples. To ensure numerical stability, we parameterize the variance as
\begin{equation}
\sigma_d^2=\log(1+\exp(\beta_d))+\epsilon.
\label{eq:variance_param}
\end{equation}
where $\beta_d$ is a learnable scalar and $\epsilon>0$ is a small constant.

In standard VID, the teacher feature and the student feature usually have the same dimensionality, so the student feature can be matched to the teacher feature directly. In our setting, however, $\widehat{X} \in \mathbb{R}^{N \times D}$ is a high-dimensional statistic, while $Z \in \mathbb{R}^{\mathbf{M \times D}}$ is a compact representation ($M \ll N$). We therefore parameterize the mean function $\mu(Z,T)\in\mathbb{R}^{N\times D}$ with a decoder that maps the compact representation back to the predictive-statistic space. 

\begin{figure*}[t]
  \vskip 0.2in
  \centering
    \centerline{\includegraphics[width=\textwidth]{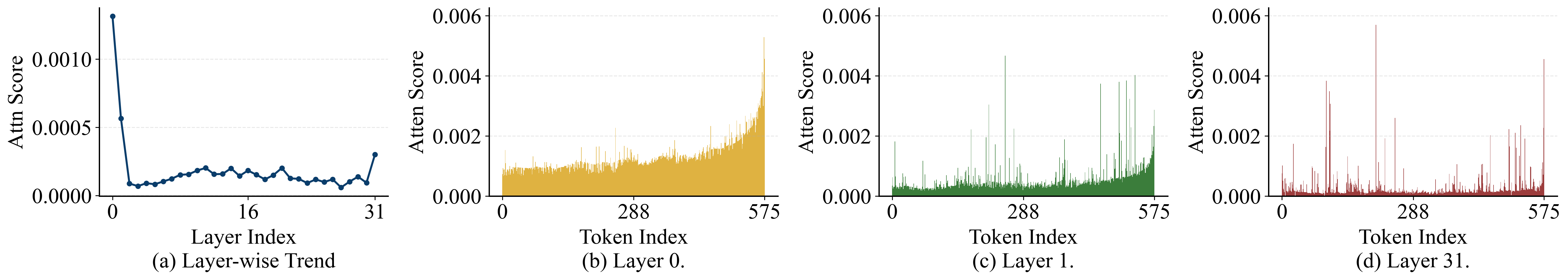}}
    \caption{
      Text-to-image cross-attention patterns in LLaVA-1.5-7B. (a) Layer-wise average of cross-modal attention scores across the LLM backbone, illustrating the concentration of interaction at initial and terminal layers. (b)--(d) Token-wise distribution of attention scores across 576 visual tokens at layers 0, 1, and 31, respectively. Higher magnitudes indicate visual regions with stronger alignment to the text instructions.
    }
    \label{attentionfigure}
\end{figure*}

\begin{table*}[t]
  \caption{Performance of ETC with different compressed-token budgets on LLaVA-1.5-7B.}
  \label{tab:llava}
  \centering
  \small
  \newcolumntype{Y}{>{\centering\arraybackslash}X}
  \begin{tabularx}{\textwidth}{l c Y Y Y Y Y Y}
    \toprule
    Methods & Tokens & MMBench & MME & SEED & SQA & VQAv2 & Avg(\%) \\
    \midrule
    LLaVA-1.5-7B & 576 & 64.00 & 1869.08 & 57.90 & 69.56 & 77.70 & 100 \\
    \midrule
    MQT-LLaVA~\cite{hu2024matryoshka} & 4 & 56.50 & 1176.10 & -- & 65.10 & 64.10 & -- \\
    LLaMA-VID~\cite{li2024llama} & 4 & -- & -- & -- & 68.70 & -- & -- \\
    VoCo~\cite{ye2025voco} & 4 & 60.40 & -- & 56.00 & -- & \textbf{74.50} & -- \\
    \rowcolor{gray!15} ETC (Ours) & 4 & \textbf{60.99} & \textbf{1668.84} & \textbf{56.66} & \textbf{68.76} & 74.10 & 95.33 \\
    \midrule
    MQT-LLaVA & 2 & 54.40 & 1144.00 & -- & 65.00 & 61.00 & -- \\
    LLaMA-VID & 2 & -- & -- & -- & \textbf{68.80} & -- & -- \\
    VoCo & 2 & 60.10 & -- & 55.00 & -- & \textbf{73.50} & -- \\
    \rowcolor{gray!15} ETC & 2 & \textbf{61.00} & \textbf{1666.84} & \textbf{55.76} & 68.47 & 72.68 & 94.55 \\
    \midrule
    QueCC~\cite{li2025inference} & 1 & 59.40 & 1269.10 & -- & \textbf{69.90} & 67.30 & -- \\
    VoCo & 1 & 58.80 & 1323.30 & 53.70 & 65.40 & 72.30 & -- \\
    \rowcolor{gray!15} ETC (Ours) & 1 & \textbf{60.22} & \textbf{1553.82} & \textbf{54.18} & 67.63 & \textbf{72.31} & 92.22 \\
    \midrule
    LLaVA-1.5-7B & 0 & 21.0 & 697.8 & 29.2 & 63.6 & 41.0 & - \\
    \bottomrule
  \end{tabularx}
\end{table*}

\subsection{Training Objective}

The final objective combines cross-entropy task loss with the VID loss for predictive-statistic preservation:
\begin{equation}
\mathcal{L}
=
\mathcal{L}_{\mathrm{CE}}
+
\lambda \mathcal{L}_{\mathrm{ETC}}(\widehat{X},Z,T),
\label{eq:final_objective}
\end{equation}
where $\mathcal{L}_{\mathrm{CE}}$ is the autoregressive cross-entropy task loss and $\lambda>0$ controls the strength of predictive-statistic preservation. In this form, the method follows the theory directly: $\mathcal{L}_{\mathrm{CE}}$ optimizes the observable task loss, cross-attention approximates the instruction-aware predictive statistic $\widehat{X}$, and $\mathcal{L}_{\mathrm{ETC}}$ encourages the compressed representation to preserve it by VID.

\newcolumntype{W}{>{\centering\arraybackslash}p{2cm}}

\begin{table*}[t]
  \caption{Performance of ETC with different compressed-token budgets on Qwen3-VL-2B.}
  \label{tab:qwen}
  \centering
  \small
  \setlength{\tabcolsep}{3pt}
  \begin{tabularx}{\textwidth}{
    l 
    c 
    >{\centering\arraybackslash}X
    W
    W
    *{4}{>{\centering\arraybackslash}X}}
    \toprule
    Method & Tokens & SQA & MMBench-CN & MMBench-EN & MME & SEED & QBench & Avg \\
    \midrule
    Qwen3VL-2B-SFT & - & 86.81 & 81.52 & 84.29 & 2087.55 & 76.84 & 60.60 & 100\% \\
    \midrule
    LLaMA-VID & 4 & 53.69 & 25.92 & 28.83 & 263.30 & 14.11 & 39.60 & 37.36 \\
    QueCC & 4 & 28.84 & 25.29 & 25.80 & 642.53 & 24.86 & 40.50 & 37.47 \\
    VoCo & 4 & 84.38 & 79.63 & 82.56 & 1413.28 & 71.10 & 55.40 & 90.75 \\
    \rowcolor{gray!15} ETC (Ours) & 4 & \textbf{85.48} & \textbf{81.44} & \textbf{82.75} & \textbf{1840.46} & \textbf{71.36} & \textbf{58.49} & \textbf{95.68} \\
    \midrule
    LLaMA-VID & 2 & 52.94 & 27.95 & 27.95 & 262.94 & 13.84 & 38.70 & 37.15 \\
    VoCo & 2 & 83.63 & 79.83 & 82.01 & 1360.18 & 69.84 & 51.80 & 88.85 \\
    \rowcolor{gray!15} ETC (Ours) & 2 & \textbf{85.23} & \textbf{80.49} & \textbf{82.72} & \textbf{1840.24} & \textbf{70.42} & \textbf{54.11} & \textbf{94.02} \\
    \midrule
    QueCC & 1 & 30.00 & 24.95 & 25.69 & 645.47 & 24.74 & 39.90 & 37.43 \\
    VoCo & 1 & 82.94 & 78.91 & 81.34 & 1333.89 & 66.74 & 49.70 & 86.93 \\
    \rowcolor{gray!15} ETC (Ours) & 1 & \textbf{84.43} & \textbf{80.39} & \textbf{82.40} & \textbf{1838.02} & \textbf{68.07} & \textbf{53.70} & \textbf{93.15} \\
    \midrule
    Qwen3VL-2B-SFT & 0 & 67.37 & 35.09 & 35.74 & 860.46 & 36.81 & 47.50 & - \\
    \bottomrule
  \end{tabularx}
\end{table*}

\subsection{Compression Architecture}

We implement the proposed compressor (Eq.~\eqref{eq:compressor}) within a decoder-only VLM architecture. This design replaces the original visual sequence $V$ with a compact representation $Z$ while maintaining the instruction-aware predictive statistics required by Eq.~\eqref{eq:psc_practical} during inference. Figure~\ref{method} illustrates the complete pipeline.

We introduce $M$ learnable compressed tokens $Z\in\mathbb{R}^{M\times d}$, where $M\ll N_v$, and construct the input sequence as
${X}=[V;Z;T]$,
where $V$ and $T$ denote visual and text tokens, respectively. The compressed tokens provide the token-level representation of $Z$. Following VoCo~\cite{ye2025voco}, we use a bottleneck attention mask to regulate information flow. The compressed tokens $Z$ attend to the visual tokens $V$, while the text tokens $T$ attend only to $Z$. This design distills instruction-aware visual information into $Z$.

Supervision is applied at the final layer of the LLM backbone. This choice is motivated by three considerations. First, the final layer is the stage most directly tied to the model output. Second, text-to-image attention in VLMs tends to peak at the initial and final layers, as illustrated in Figure~\ref{attentionfigure}, indicating that these layers are critical position for cross-modal interaction. Third, compared with early layers that primarily aggregate local features, the final layer contains more mature multimodal semantics and exhibits stronger sparsity over task-relevant tokens. Aligning features at final layer therefore facilitates the extraction of high-density, query-relevant information.

At the final layer, we compute the practical predictive-statistic target in Eq.~\eqref{eq:xw_def} using the original visual features. Specifically, text-to-image cross-attention provides instruction-aware weights, which reweight the final-layer visual representation to obtain $\widehat{X}$. Since $Z$ contains only $M$ tokens whereas $X_v$ contains $N_v$ tokens, an MLP decoder maps $Z$ back to the original visual token dimension. The decoded representation parameterizes the mean function $\mu(Z,T)$ of the variational distribution in Eq.~\eqref{eq:gaussian_decoder}.

During training, $\mathcal{L}_{\mathrm{ETC}}$ is optimized together with the autoregressive cross-entropy loss $\mathcal{L}_{\mathrm{CE}}$ in Eq.~\eqref{eq:final_objective}. Thus, the model learns both to predict downstream outputs and to preserve in $Z$ the instruction-aware visual information needed for prediction.

During inference, $Z$ is obtained in the prefilling stage by applying the same bottleneck mask, effectively compressing the information in $V$ into $Z$. The original visual tokens $V$ are then discarded from the KV cache. In the subsequent decoding stage, generation is performed using only $Z$ and the text tokens $T$. This retains instruction-relevant visual information while reducing decoding cost and improving inference efficiency.

\begin{figure}[t]
  \centering
  \includegraphics[width=0.9\linewidth]{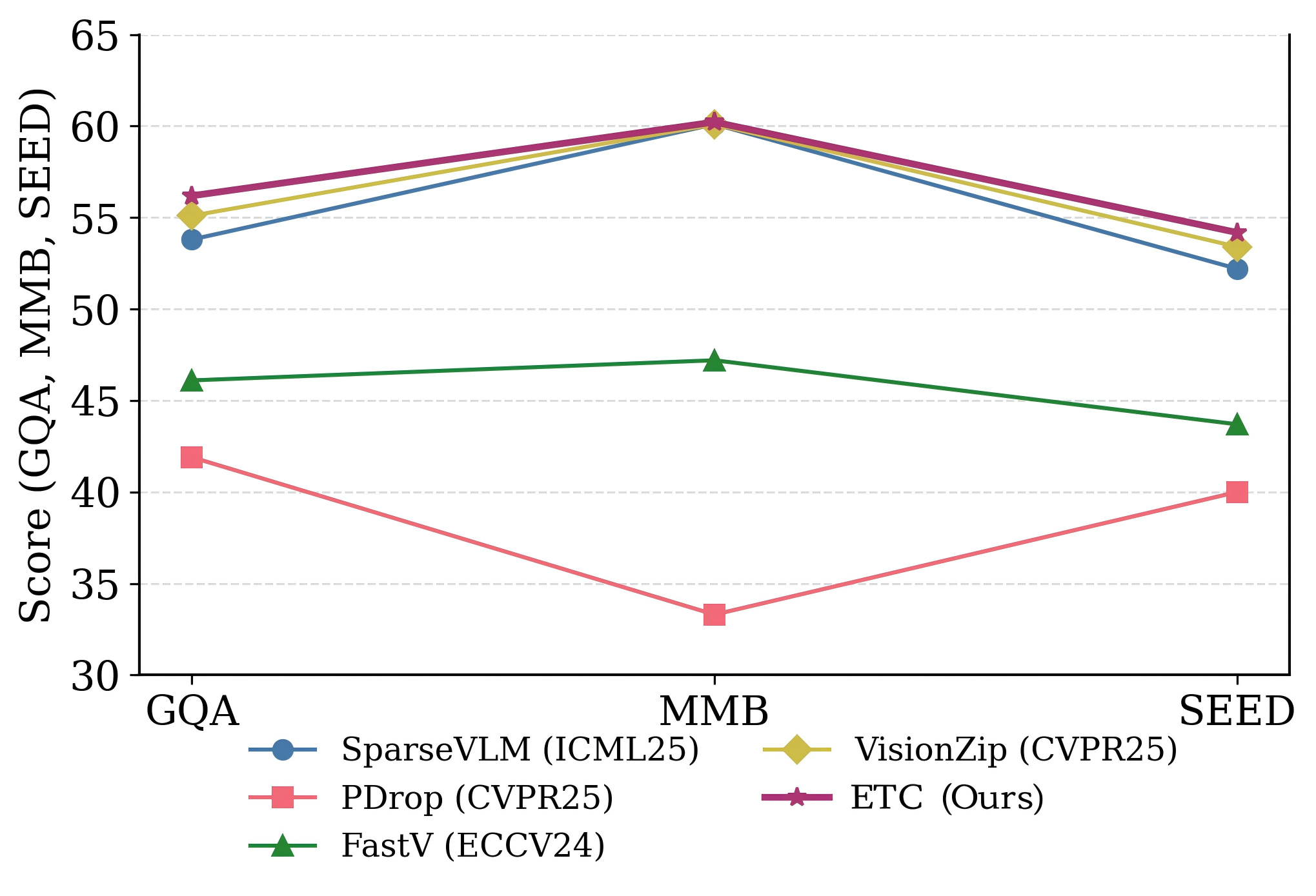}
  \caption{Comparison between selective compression methods and ETC.}
  \label{Selective}
\end{figure}

\section{Experiments}
\label{sec:experiments}
\subsection{Experimental Setup}
\textbf{Datasets and Models.} We perform Supervised Fine-Tuning (SFT) on the LLaVA-v1.5-mix665k instruction-tuning dataset, which combines COCO~\cite{lin2014microsoft}, GQA~\cite{hudson2019gqa}, OCR-VQA~\cite{mishra2019ocr}, TextVQA~\cite{singh2019towards}, and Visual Genome~\cite{krishna2017visual}. We evaluate two representative architectures: LLaVA-1.5-7B~\cite{liu2024improved}, a widely used model for token compression, and Qwen3-VL-2B~\cite{yang2025qwen3}, a strong open-source VLM with native dynamic resolution and 3D rotary positional embeddings.

\textbf{Benchmarks.} We evaluate on MMBench~\cite{liu2024mmbench}, MME~\cite{fu2025mme}, SEED-Bench~\cite{li2024seed}, ScienceQA~\cite{lu2022learn}, VQAv2~\cite{goyal2017making}, and Q-Bench~\cite{wu2023q}.

\columnbreak

\begin{table}[t]
  \centering
  \caption{RefCOCO results of ETC and other compression methods.}
  \label{tab:etc_refcoco}
  \small
  \begin{tabularx}{\linewidth}{l 
    >{\centering\arraybackslash}X 
    >{\centering\arraybackslash}X 
    >{\centering\arraybackslash}X}
    \toprule
    \multirow{2}{*}{Method} & \multirow{2}{*}{Tokens} & \multicolumn{2}{c}{RefCOCO bbox} \\
    \cmidrule(lr){3-4}
    & & TestA & TestB \\
    \midrule
    Qwen3VL-2B-SFT & - & 20.78 & 41.48 \\
    \midrule
    QueCC & 1 & 16.31 & 33.96 \\
    VoCo & 1 & 10.38 & 28.51 \\
    ETC (Ours) & 1 & \textbf{18.73} & \textbf{37.72} \\
    \bottomrule
  \end{tabularx}
\end{table}

\begin{table*}[t]
  \caption{Efficiency analysis of ETC, including KV-cache memory, CUDA time, and FLOPs. $\Delta$ denotes the reduction ratio.}
  \label{tab:time}
  \centering
  \small
  \begin{tabular*}{\linewidth}{@{\extracolsep{\fill}} lccccccc @{}}
    \toprule
    Method & Image Tokens & KV cache (MB) & $\Delta$ & CUDA Time (ms) & $\Delta$ & FLOPs (T) & $\Delta$ \\
    \midrule
    Full cache & 1024 & 223.54 & - & 203.05& - & 2.98 & - \\
    ETC & 1  & 10.44 & 95.33 & 114.13 & 43.79 & 1.37 &54.03 \\
    \bottomrule
  \end{tabular*}
\end{table*}

\begin{table*}[htbp]
  \centering
  \begin{minipage}[t]{0.49\textwidth}
    \centering
    \captionof{table}{Ablation of the instruction-aware mixing coefficient $\alpha$ across SQA, MMBench-CN, MMBench-EN, and QBench.}
    \label{alphaattenntion}
    \small
    \setlength{\tabcolsep}{4pt}
    \begin{tabularx}{\linewidth}{l *{4}{>{\centering\arraybackslash}X}}
      \toprule
      $\alpha$ & SQA & MMBench-CN & MMBench-EN & QBench \\
      \midrule
      0    & 82.59 & 79.6  & 81.64 & 51.5 \\
      0.10 & 82.64 & 79.6  & 81.7  & 52.5 \\
      0.20 & 83.59 & 78.56 & 81.5  & 52.7 \\
      0.40 & 83.14 & 79.26 & 81.82 & 52.4 \\
      0.60 & \textbf{84.43} & \textbf{80.39} & \textbf{82.4} & \textbf{53.7} \\
      0.80 & 82.69 & 79.93 & 82.24 & 51.4 \\
      1.00 & 82.89 & 79.89 & 82.05 & 52.2 \\
      \bottomrule
    \end{tabularx}
  \end{minipage}\hfill
  \begin{minipage}[t]{0.47\textwidth}
    \centering
    \captionof{table}{Ablation of the ETC loss weight $\lambda$ across SQA, MMBench-CN, MMBench-EN, and QBench.}
    \label{tab:lambda_ablation}
    \small
    \setlength{\tabcolsep}{4pt}
    \begin{tabularx}{\linewidth}{l *{4}{>{\centering\arraybackslash}X}}
      \toprule
      $\lambda$ & SQA & MMBench-CN & MMBench-EN & QBench \\
      \midrule
      0        & 82.94 & 78.91 & 81.34 & 49.70 \\
      $10^{-5}$ & \textbf{84.43} & \textbf{80.39} & \textbf{82.4} & \textbf{53.7} \\
      $10^{-4}$ & 82.94 & 79.51 & 81.89 & 52.8 \\
      $10^{-3}$ & 82.59 & 79.83 & 81.75 & 52.2 \\
      $10^{-2}$ & 81.6  & 79.28 & 81.38 & 51.4 \\
      $10^{-1}$ & 82.84 & 79.16 & 81.08 & 51.9 \\
      $10^{0}$  & 84.03 & 79.65 & 82.12 & 50.7 \\
      \bottomrule
    \end{tabularx}
  \end{minipage}
\end{table*}

\begin{table}[t]
	\centering
	\caption{Comparison of alignment strategies across SQA, MME, and QBench.}
	\label{tab:align_compare}
	\small
	\setlength{\tabcolsep}{4pt}
	\begin{tabularx}{\columnwidth}{l *{3}{>{\centering\arraybackslash}X}}
		\toprule
		Method & SQA (\%) & MME & QBench (\%) \\
		\midrule
		None & 82.94 & 1333.89 & 49.70 \\
		MSE & 84.13 & 1741.05 & 50.50 \\
		VID (ETC) & \textbf{84.43} & \textbf{1838.02} & \textbf{53.70} \\
		\bottomrule
	\end{tabularx}
\end{table}

\textbf{Implementation Details.} We freeze the vision tower and optimize the cross-modal projector and LLM backbone with AdamW on NVIDIA GeForce RTX 3090 GPUs. The model is trained with the joint objective in Eq.~\eqref{eq:final_objective}, using $\lambda = 10^{-5}$ for the ETC loss and $\alpha = 0.6$ for instruction-aware approximation. To accommodate the 3D rotary positional embedding in Qwen3-VL, we specifically adapt the indexing for the $M$ compressed tokens. The temporal index for all compressed tokens is set to $T_{visual} + 1$. Regarding spatial indices, the first compressed token's $W$ and $H$ are aligned with its $T$, while subsequent tokens' spatial indices are incremented sequentially to maintain unique positional identity.

\subsection{Experimental Results}
\textbf{Performance on LLaVA-1.5-7B.} Table~\ref{tab:llava} compares ETC with existing abstractive compression methods on LLaVA-1.5-7B. At a 4-token budget, ETC achieves the best results on MMBench (60.99\%), MME (1668.84), SEED (56.66\%), and SQA (68.76\%), outperforming MQT-LLaVA by 492.74 points on MME. Under extreme compression at a single token, ETC still achieves the best performance on four metrics, including 60.22\% on MMBench and 1553.82 on MME. The performance remains robust across token budgets of 4, 2, and 1.

\textbf{Performance on Qwen3-VL-2B.} To further validate scalability across architectures, we evaluate ETC on Qwen3-VL-2B. Table~\ref{tab:qwen} shows that, at a 4-token budget, ETC achieves the best results on all metrics, including 85.48\% on SQA and 1840.46 on MME, while retaining 95.68\% of the performance of the Qwen3VL-2B-SFT baseline on average. Compared with VoCo, ETC improves MME by 427.18 points and QBench by 3.09 points. This trend remains consistent in the 1-token setting: ETC retains 93.15\% of the baseline performance on average, outperforming VoCo (86.93\%) and QueCC (37.43\%). These results suggest stable behavior across both LLaVA and Qwen backbones.

\textbf{Comparisons with Selective Compression Methods.} Figure~\ref{Selective} compares ETC with selective compression methods, including SparseVLM\cite{zhang2025sparsevlm}, VisionZip\cite{yang2025visionzip}, PDrop\cite{11094455}, and FastV\cite{chen2024image}. ETC compresses visual tokens into a single token, while other methods compress them into 64 tokens. Using a single token, ETC achieves 56.19\% on GQA, 60.22\% on MMBench, and 54.18\% on SEED, outperforming all 64-token methods on these benchmarks. These results suggest that ETC preserves task-relevant information even under extreme compression.

\textbf{RefCOCO Results.} We further evaluate ETC on RefCOCO to assess fine-grained region-text alignment under compression. As shown in Table~\ref{tab:etc_refcoco}, ETC is the best-performing compressed method, reaching 18.73 on TestA and 37.72 on TestB. This suggests that ETC preserves task-relevant visual evidence more effectively than other compression methods. Overall, performance remains limited may because the 2B backbone is relatively weak, and bounding-box tasks have high information density, requiring more tokens. Further analysis of task difficulty and token budgets is provided in Appendix~\ref{app:token_budget_analysis}.

\textbf{Efficiency Analysis.} Table~\ref{tab:time} shows that, relative to a full cache of $1{,}024$ visual tokens, ETC reduces the KV cache from $223.54$ MB to $10.44$ MB ($95.33\%$), CUDA time from $203.05$ ms to $114.13$ ms, and theoretical FLOPs from $2.98$ T to $1.37$ T, substantially lowering visual-stream inference cost.

\section{Discussion}
\label{sec:discussion}
\subsection{Sensitivity to Alpha and Lambda}
We examine the two scalar coefficients involved in ETC: the instruction-aware mixing coefficient $\alpha$ in Equation~\ref{eq:xw_def} and the ETC loss weight $\lambda$ in Equation~\ref{eq:final_objective}. As shown in Table~\ref{alphaattenntion}, performance is lowest in the task-agnostic setting ($\alpha=0$), where the model yields 82.59 on SQA, 79.6 on MMBench-CN, 81.64 on MMBench-EN, and 51.5 on QBench. In this case, all visual patches are treated uniformly, making it harder for the compressed representation to concentrate on instruction-relevant regions. As $\alpha$ increases, the query is used as a relevance signal to emphasize more task-relevant visual content, and performance peaks at $\alpha=0.60$, where ETC reaches 84.43 on SQA, 80.39 on MMBench-CN, 82.4 on MMBench-EN, and 53.7 on QBench. When $\alpha$ is increased further to 1.00, the scores decline to 82.89, 79.49, 82.05, and 52.2, respectively, suggesting that overly aggressive weighting may suppress auxiliary cues that are still useful for reasoning.

Table~\ref{tab:lambda_ablation} shows a similar trend for $\lambda$. Without the ETC term ($\lambda=0$), the model obtains 82.94 on SQA, 78.91 on MMBench-CN, 81.34 on MMBench-EN, and 49.70 on QBench. Introducing a small positive weight substantially improves all four benchmarks, with the best overall result achieved at $\lambda=10^{-5}$. Over the broader range from $10^{-4}$ to $10^{0}$, the performance changes remain moderate, indicating that ETC loss weight $\lambda$ is not overly sensitive.

\subsection{Validity of the Variational Information Distillation Objective}

To validate the second component of ETC, we evaluate three variants regarding the preservation of the instruction-aware predictive-statistic estimate $\widehat{X}$: no auxiliary loss, direct MSE regression, and the proposed variational information distillation objective. Table~\ref{tab:align_compare} shows that VID performs best on all three benchmarks, with especially clear gains over MSE on MME and QBench.

This result is expected under extreme compression. MSE enforces point-wise Euclidean alignment, which tends to pull the compressed token toward an averaged representation when a single token must summarize many visual tokens, causing a mean-collapse effect that blurs fine-grained semantics. In contrast, VID optimizes a tractable variational bound on the mutual information between $\widehat{X}$ and $(Z,T)$, while using a learnable variance to accommodate reconstruction uncertainty. Therefore, ETC encourages $\widehat{X}$ to remain recoverable from the compressed representation rather than merely close in $\ell_2$ distance, which is more suitable for extreme token compression.

\begin{table}[t]
  \centering
  \caption{Comparison of supervision layers.}
  \label{tab:layer_performance}
  \small
  \setlength{\tabcolsep}{4pt}
  \begin{tabular*}{\linewidth}{@{\extracolsep{\fill}}cccc@{}}
    \toprule
    Layer & SQA(\%) & MMBench-EN(\%) & MME \\
    \midrule
    0  & 83.19 & 81.94 & 1,796.61 \\
    15 & 83.34 & 82.05 & 1,755.84 \\
    32 & \textbf{84.43} & \textbf{82.40} & \textbf{1,838.02} \\
    \bottomrule
  \end{tabular*}
\end{table}

\subsection{Layer-wise Alignment Analysis}

We ablate where ETC supervision is applied when constructing $\widehat{X}$ and optimizing $\mathcal{L}_{\mathrm{ETC}}$. As shown in Table~\ref{tab:layer_performance}, the final layer ($L=32$) performs best across all reported benchmarks, improving SQA from 83.19 to 84.43 and MMBench-EN from 81.94 to 82.40 compared with $L=0$, while also achieving the highest MME score (1838.02). This supports the method design: the final layer is most directly tied to prediction and contains more task-relevant multimodal semantics, making it the most suitable layer for predictive-statistic preservation.

\begin{table}[t]
  \centering
  \caption{Comparison of decoder parameterizations.}
  \label{tab:model_performance}
  \small
  \setlength{\tabcolsep}{3pt}
  \begin{tabular*}{\linewidth}{@{\extracolsep{\fill}}lccc@{}}
    \toprule
    Model & SQA(\%) & MMBench-EN(\%) & MME \\
    \midrule
    Q-Former & 83.04 & 81.96 & 1,822.15 \\
    linear interpolation & 84.38 & 81.64 & 1,744.50 \\
    MLP & \textbf{84.43} & \textbf{82.40} & \textbf{1,838.02} \\
    \bottomrule
  \end{tabular*}
\end{table}

\subsection{Decoder Parameterizations}

We compare three choices for parameterizing the mean function $\mu(Z,T)$ of the variational distribution in Eq.~\eqref{eq:gaussian_decoder}: an MLP decoder~\cite{liu2024improved}, QFormer~\cite{li2023blip}, and linear interpolation~\cite{chen2023extending}. Table~\ref{tab:model_performance} shows that the MLP decoder achieves the best overall results, reaching 1838.02 on MME and 82.40 on MMBench-EN. This indicates that a simple non-linear decoder is sufficient to recover the target $\widehat{X}$ from compressed tokens, whereas linear interpolation underfits and Q-Former offers no clear gain, potentially because the powerful downstream decoder alleviates the learning pressure on the compression module, leading to suboptimal representation alignment. We therefore use the MLP decoder as the default parameterization of $\mu(Z,T)$.

\bibliographystyle{unsrtnat}
\bibliography{ms}

\clearpage
\appendix
\section{Appendix}
\subsection{Proofs and Additional Derivations}
\label{app:proofs}
This appendix provides the full derivations for Section~\ref{sec:method}. We derive the ideal requirement for the compact representation $Z$ from task-loss minimization, and then approximate the practical objective based on cross-attention and variational information distillation.

\subsubsection{Conditions for Minimizing Task Loss}
\label{app:recoverable_statistic}

Let $V \in \mathcal{V}$ denote the original visual token sequence, $T \in \mathcal{T}$ the text instruction or query, $Y \in \mathcal{Y}$ the target output sequence, and $Z\in\mathcal{Z}$ the compact representation produced by the compressor $f_\theta$:
\begin{equation}
Z=f_\theta(V,T).
\label{eq:compressor_def}
\end{equation}

Under the standard negative log-likelihood objective, the conditional Bayes risk under log-loss with an representation $S$ is
\begin{equation}
\mathcal{R}^*(S)=\min_q \mathbb{E}\big[-\log q(Y\mid S,T)\big]=H(Y\mid S,T),
\end{equation}
where $q(Y\mid S,T)$ is the predictive distribution conditioned on $S$ and $T$, and $H(\cdot\mid\cdot)$ denotes conditional entropy.

Applying this identity to the full visual input $V$ and the compressed representation $Z$ gives
\begin{equation}
\Delta_{\mathrm{task}}(Z)
:=
\mathcal{R}^*(Z)-\mathcal{R}^*(V)
=
H(Y\mid Z,T)-H(Y\mid V,T).
\label{eq:task_gap_appendix}
\end{equation}
Here $\Delta_{\mathrm{task}}(Z)$ is the excess task loss incurred when $Z$ replaces $V$ for downstream prediction.

Eq.~\eqref{eq:compressor_def} implies that Z introduces no additional information beyond $(V,T)$. Therefore,
\begin{equation}
H(Y\mid V,T)=H(Y\mid V,Z,T).
\end{equation}
Substituting this identity into the Eq.~\eqref{eq:task_gap_appendix}
\begin{equation}
\Delta_{\mathrm{task}}(Z)
=
H(Y\mid Z,T)-H(Y\mid V,Z,T)
=
I(Y;V\mid Z,T),
\end{equation}
where $I(\cdot;\cdot\mid\cdot)$ denotes conditional mutual information.

To relate the task loss to the information preserved in $Z$, expand the joint conditional mutual information $I(Y;V,Z\mid T)$ in two ways:
\begin{equation}
I(Y;V,Z\mid T)=I(Y;Z\mid T)+I(Y;V\mid Z,T).
\end{equation}
and
\begin{equation}
I(Y;V,Z\mid T)=I(Y;V\mid T)+I(Y;Z\mid V,T).
\end{equation}
Since $Z=f_\theta(V,T)$ is fully determined by $(V,T)$, we have $I(Y;Z\mid V,T)=0$. Therefore,
\begin{equation}
\Delta_{\mathrm{task}}(Z)=I(Y;V\mid Z,T)=I(Y;V\mid T)-I(Y;Z\mid T).
\end{equation}
This shows that minimizing the task loss is equivalent to minimizing the task-relevant information discarded by compression, namely $I(Y;V\mid Z,T)$, or equivalently maximizing the task-relevant information retained in $Z$, namely $I(Y;Z\mid T)$.

The ideal zero-loss limit satisfies
\begin{equation}
\Delta_{\mathrm{task}}(Z)=0
\iff
I(Y;V\mid Z,T)=0
\iff
Y \perp V \mid (Z,T),
\end{equation}
which means that once the compact representation $Z$ and the instruction $T$ are given, the original visual input $V$ provides no additional predictive information about $Y$. $Z$ is a sufficient statistic of $V$ for predicting $Y$ conditioned on the instruction $T$.

To express the ideal predictive sufficient statistic, define the latent statistic
\begin{equation}
X^\star=g^\star(V,T):=p(\cdot\mid V,T).
\end{equation}
Here $g^\star$ maps the pair $(V,T)$ to the predictive sufficient statistic, $X^\star$ denotes the ideal instruction-aware predictive statistic in the uncompressed input, and $p(\cdot\mid \cdot,\cdot)$ denotes the conditional distribution.

Again because $Z=f_\theta(V,T)$, conditioning on $(V,T)$ or on $(V,Z,T)$ gives the same conditional law:
\begin{equation}
p(\cdot\mid V,T)=p(\cdot\mid V,Z,T).
\end{equation}
Under the zero-loss condition $Y \perp V \mid (Z,T)$, we also have
\begin{equation}
p(\cdot\mid V,Z,T)=p(\cdot\mid Z,T).
\end{equation}
Combining the last two equalities yields
\begin{equation}
X^\star=p(\cdot\mid V,T)=p(\cdot\mid Z,T).
\end{equation}
If we define $h^\star(Z,T):=p(\cdot\mid Z,T)$, where $h^\star$ maps the compressed representation and the instruction to the same predictive distribution, then
\begin{equation}
X^\star=h^\star(Z,T).
\end{equation}
Hence $X^\star$ is measurable with respect to $(Z,T)$, and therefore
\begin{equation}
H(X^\star\mid Z,T)=0.
\end{equation}
Here $H(X^\star\mid Z,T)$ is the conditional entropy of the latent predictive statistic after observing $Z$ and $T$. The value zero means that the ideal predictive statistic can be recovered exactly from the compact representation and the instruction.

\subsubsection{Why Cross-Attention Provides a Practical Proxy}

For $j^{th}$ text token, let $A_{j,i}$ denote the text-to-image attention score assigned to $i^{th}$ visual token, let $W_V$ be the value-weight matrix in cross-attention, and let $o_j\in\mathbb{R}^{D}$ be the visual contribution to the cross-attention output for text token. Then
\begin{equation}
o_j=\sum_{i=1}^{N_v} A_{j,i}W_Vx_i.
\end{equation}
Let $o_j^{(-i)}\in\mathbb{R}^{D}$ denote the output after removing $i^{th}$ visual token while keeping all attention weights fixed. Then
\begin{equation}
o_j^{(-i)}=\sum_{k\neq i} A_{j,k}W_Vx_k.
\end{equation}
Subtracting the two expressions gives
\begin{equation}
o_j-o_j^{(-i)}=A_{j,i}W_Vx_i.
\label{eq:fixed_ablation_delta}
\end{equation}
Taking norms yields
\begin{equation}
\|o_j-o_j^{(-i)}\|\le A_{j,i}\|W_Vx_i\|,
\label{eq:attn_bound_single}
\end{equation}
because attention weights are non-negative. Summing over all text positions gives
\begin{equation}
\sum_{j=1}^{N_t}\|o_j-o_j^{(-i)}\|
\le
\|W_Vx_i\|\sum_{j=1}^{N_t}A_{j,i}.
\label{eq:attn_bound_total}
\end{equation}

Equation~\eqref{eq:attn_bound_total} shows that, under fixed-attention ablation, the change in the text-side representation caused by removing the $i^{th}$ visual token is upper bounded by the product of its attention score and the norm of its value vector. Thus, larger aggregated attention score indicates that a visual token plays a stronger role in shaping the downstream text representation.

This analysis explains why aggregated cross-attention scores can serve as a proxy for instruction-conditioned predictive relevance. However, it does not establish exact causal attribution under full recomputation of attention and hidden states.

\subsubsection{VID Objective for Preserving \texorpdfstring{$\widehat{X}$}{X-hat}}

Once $\widehat{X}$ is defined, the second part of Eq.~\eqref{eq:psc_practical} requires it to remain recoverable from $(Z,T)$. The standard VID~\cite{ahn2019variational} gives a tractable objective for this requirement.

Starting from the definition of conditional mutual information,
\begin{equation}
\begin{aligned}
I(\widehat{X};Z\mid T)&=H(\widehat{X}\mid T)-H(\widehat{X}\mid Z,T) \\
&=H(\widehat{X}\mid T)+\mathbb{E}_{p(\widehat{X},Z,T)}\big[\log p(\widehat{X}\mid Z,T)\big],
\end{aligned}
\end{equation}
where $\widehat{X}$ is the practical predictive-statistic proxy, $p(\widehat{X}\mid Z,T)$ is the true conditional distribution of $\widehat{X}$ given $(Z,T)$, and the expectation is taken over the joint distribution of $(\widehat{X},Z,T)$. We introduce a variational distribution $q(\widehat{X}\mid Z,T)$ to approximate $p(\widehat{X}\mid Z,T)$ This gives
\begin{equation}
\begin{aligned}
I(\widehat{X};Z\mid T)
= {} & H(\widehat{X}\mid T)+\mathbb{E}_{p(\widehat{X},Z,T)}\big[\log q(\widehat{X}\mid Z,T)\big] \\
& +\mathbb{E}_{p(Z,T)}\!\left[
\mathrm{KL}\!\left(
p(\widehat{X}\mid Z,T)\,\|\,q(\widehat{X}\mid Z,T)
\right)
\right],
\end{aligned}
\end{equation}
where $\mathrm{KL}(\cdot\|\cdot)$ denotes the Kullback-Leibler divergence. Because the KL term is non-negative, we obtain
\begin{equation}
I(\widehat{X};Z\mid T)
\ge
H(\widehat{X}\mid T)+\mathbb{E}_{p(\widehat{X},Z,T)}\big[\log q(\widehat{X}\mid Z,T)\big].
\end{equation}
Since $H(\widehat{X}\mid T)$ does not depend on the model parameters, maximizing $I(\widehat{X};Z\mid T)$ is equivalent to minimizing
\begin{equation}
\mathcal{L}_{\mathrm{ETC}}
=
-\mathbb{E}_{p(\widehat{X},Z,T)}\big[\log q(\widehat{X}\mid Z,T)\big].
\end{equation}

In standard VID, teacher and student features usually have matched dimensionality. In our setting, however, the proxy predictive statistic $\widehat{X}\in\mathbb{R}^{N\times D}$ and the compact representation $Z\in\mathbb{R}^{M\times D}$ are separated by a bottleneck with $M\ll N$. We therefore parameterize the mean function $\mu(Z,T)\in\mathbb{R}^{N\times D}$ with an MLP decoder and use the Gaussian variational distribution
\begin{equation}
  q(\widehat{X}\mid Z,T)
  =
  \prod_{n=1}^{N}\prod_{d=1}^{D}
  \mathcal{N}\!\big(\widehat{X}_{n,d};\mu_{n,d}(Z,T),\sigma_d^2\big),
\end{equation}
where $n\in\{1,\ldots,N\}$ indexes token positions in $\widehat{X}$, $d\in\{1,\ldots,D\}$ indexes feature dimensions, $\mu_{b,n,d}(Z,T)$ is the $(n,d)$-th entry of the decoder mean, and $\sigma_d^2$ is the learnable variance shared across batch. In practice, $N=N_v$, so the decoder output is aligned with the same final-layer visual-token space as $\widehat{X}$. Substituting this Gaussian form into Eq.~\eqref{eq:suf_general} yields
\begin{equation}
\begin{aligned}
  \mathcal{L}_{\mathrm{ETC}}
  =
  &\frac{1}{B}\sum_{b=1}^{B}\left[
  \sum_{d=1}^{D}\left(
  \frac{1}{2\sigma_d^2}\sum_{n=1}^{N}
  \big(\widehat{X}_{b,n,d}-\mu_{b,n,d}(Z,T)\big)^2
  + N\log \sigma_d
  \right)
  \right] \\
  &+ C,
\end{aligned}
\end{equation}
where $B$ is the size of batch, $b\in\{1,\ldots,B\}$ indexes training samples and $C=\frac{ND}{2}\log(2\pi)$ is a constant independent of the model parameters and can be omitted during optimization. The empirical objective becomes
\begin{equation}
\mathcal{L}_{\mathrm{ETC}}
=
\frac{1}{B}\sum_{b=1}^{B}\left[
\sum_{d=1}^{D}\left(
\frac{1}{2\sigma_d^2}\sum_{n=1}^{N}
\big(\widehat{X}_{b,n,d}-\mu_{b,n,d}(Z,T)\big)^2
+ N\log \sigma_d
\right)
\right].
\end{equation}

The variance is parameterized as
\begin{equation}
\sigma_d^2=\log(1+\exp(\beta_d))+\epsilon,
\end{equation}
where $\beta_d$ is a learnable scalar and $\epsilon>0$ is a small constant used for numerical stability. This parameterization enforces positivity and improves training stability. This completes the bridge from the ideal sufficiency condition in Eq.~\eqref{eq:ideal_psc} to the practical training objective in Eq.~\eqref{eq:suf_gaussian}.

\begin{table*}[t]
  \centering
  \caption{Hyperparameter settings for LLaVA-1.5-7B and Qwen3-VL-2B fine-tuning.}
  \label{tab:hyperparameter_comparison}
  \small
  \begin{tabularx}{\linewidth}{
    l
    X
    >{\centering\arraybackslash}X
    >{\centering\arraybackslash}X
  }
    \toprule
    Category & Hyperparameter & LLaVA-1.5-7B & Qwen3-VL-2B \\
    \midrule
    Model & Base LLM & Vicuna-7B-v1.5 & Qwen3-2B-Instruct \\
                           & Training Type & Full Fine-Tuning & LoRA \\
                           & LoRA Rank / Alpha & -- & 8 / 32 \\
                           & Precision & bfloat16 & bfloat16 \\
    \midrule
    Training & Learning Rate & 2e-5 & 1e-4 \\
                              & Optimizer & AdamW & AdamW \\
                              & Total Batch Size & 128 & 8 \\
                              & LR Scheduler & Cosine decay & Cosine decay \\
                              & Epochs & 1 & 1 \\
                              & Warmup Ratio & 0.03 & 0.05 \\
                              & Weight Decay & 0.0 & 0.0 \\
    \midrule
    ETC & Compressed Tokens (M) & 1 & 1 \\
                            & ETC Loss Weight ($\lambda$) & 1e-5 & 1e-5 \\
                            & Instruction-aware Weighting ($\alpha$) & 0.6 & 0.6 \\
                            & Alignment Layer & -1 & -1 \\
                            & ETC Rank & 8 & 8 \\
    \bottomrule
  \end{tabularx}
\end{table*}

\subsection{Additional Experimental Details}

\subsubsection{Training Datasets}
\textbf{LLaVA-v1.5-mix665k Overview.}\footnote{LLaVA instruction data file: \url{https://huggingface.co/datasets/liuhaotian/LLaVA-Instruct-150K/blob/main/llava_instruct_150k.json}. Data preparation and collection documentation: \url{https://github.com/haotian-liu/LLaVA/blob/main/docs/Data.md}.} The primary instruction-tuning source is the LLaVA-v1.5-mix665k dataset, a large-scale multimodal collection designed to enhance the model's ability to follow complex instructions. It integrates a diverse range of visual tasks by consolidating several specialized datasets, ensuring robust performance across general visual QA, spatial reasoning, and text-centric visual understanding.

\textbf{COCO} is a foundational dataset for object detection, segmentation, and captioning. It contains over 200,000 labeled images across 80 object categories. In the context of instruction tuning, it provides the model with a fundamental understanding of scene composition and the relationships between everyday objects in natural environments.

\textbf{GQA} focuses on real-world visual reasoning and compositional question answering. Unlike standard VQA, it utilizes scene graphs to create questions that require multistep reasoning (e.g., spatial relations and attribute identification). This dataset is instrumental in improving the model's logical consistency and its ability to handle complex, structured queries.

\textbf{OCR-VQA} is designed to bridge the gap between visual recognition and optical character recognition (OCR). It consists of over 200,000 images of book covers and documents associated with 1 million question-answer pairs. This dataset enables the model to read, interpret, and reason about text embedded within various visual contexts.

\textbf{TextVQA} specifically challenges models to read and reason about text found in natural images (e.g., street signs, storefronts, and labels). By requiring the model to extract text to answer questions, it significantly enhances the system's ``OCR-centric'' visual reasoning capabilities beyond structured document formats.

\textbf{Visual Genome} provides dense annotations of image components, including objects, attributes, and relationships, mapped to a formal knowledge base. It facilitates a deeper understanding of visual concepts by providing more granular detail than standard captioning datasets, allowing the model to ground its language in specific visual regions.

\begin{figure*}[htbp]
	\vskip 0.2in
	\centering
	\includegraphics[width=0.49\textwidth]{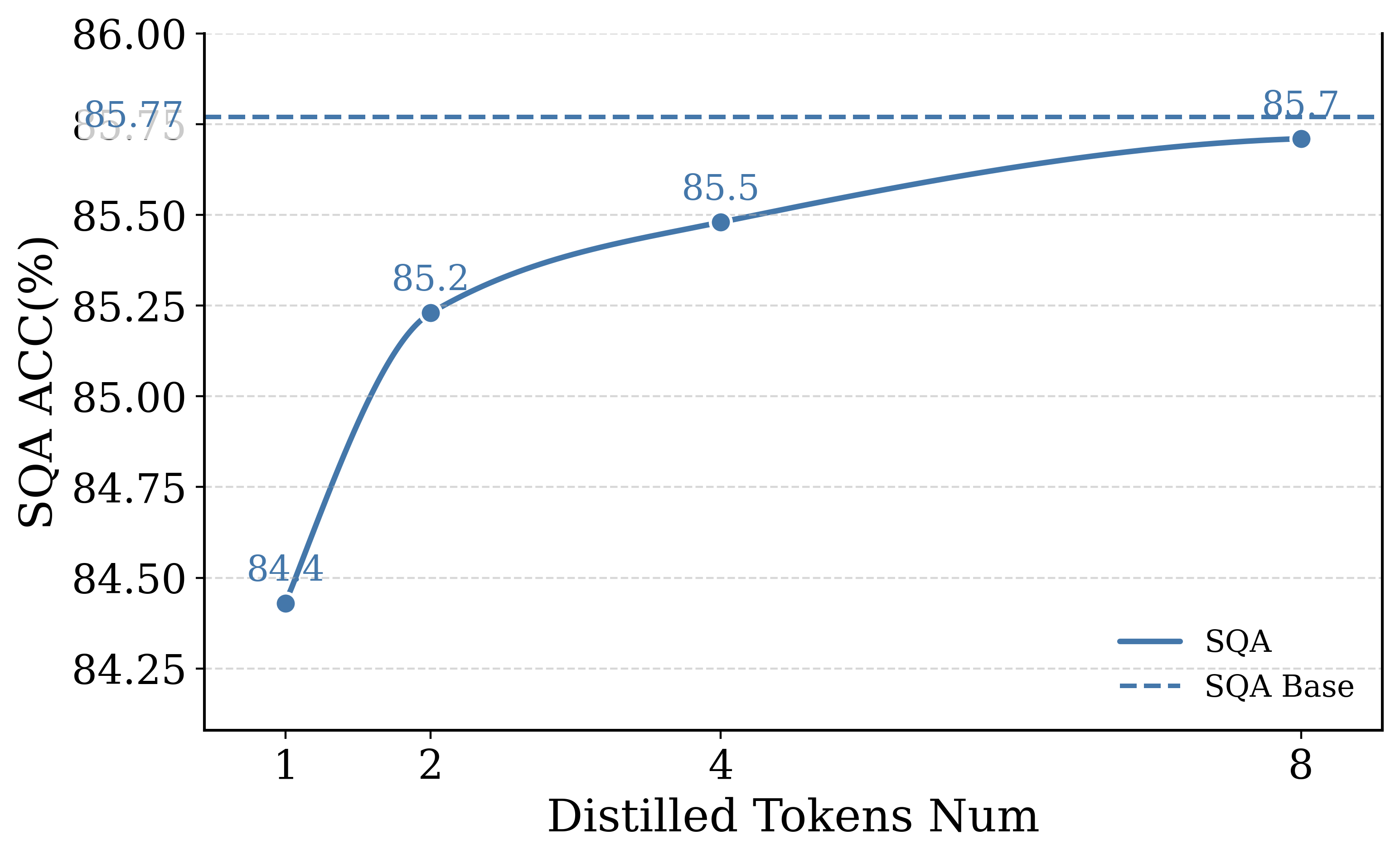}
	\includegraphics[width=0.49\textwidth]{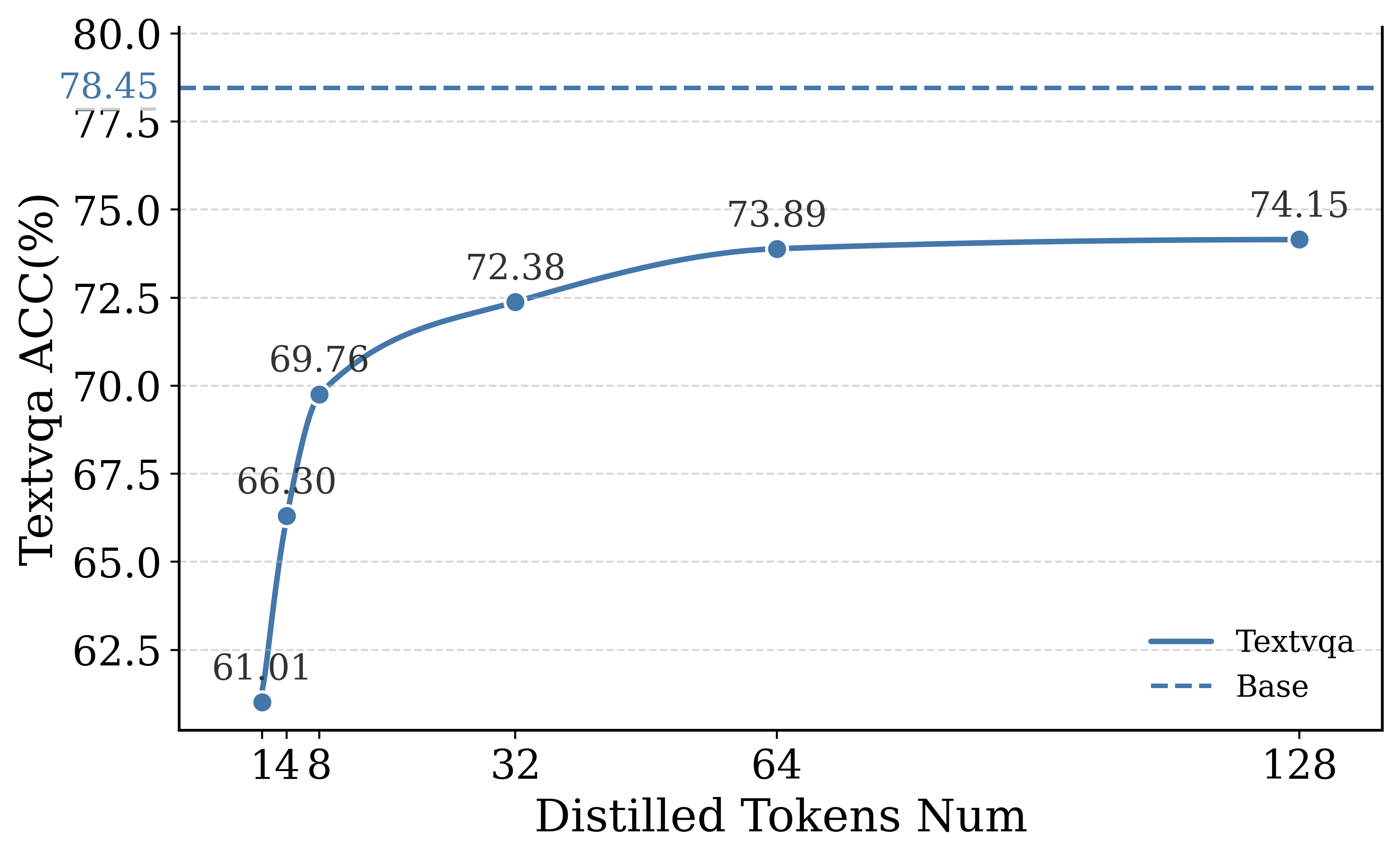}
	\caption{
		Token-budget scaling behavior of ETC on SQA (left) and TextVQA (right). ETC approaches the full-token baseline quickly on SQA, while TextVQA improves more gradually as additional compressed tokens are introduced. The dashed line in the figure represents the accuracy of the Qwen3VL-2B-SFT baseline.
	}
	\label{fig:etc_token_scaling}
\end{figure*}

\subsubsection{Benchmarks}
\textbf{MMBench} is a comprehensive evaluation pipeline that employs a ``Circular Eval'' strategy to robustly assess a model's various capabilities. It covers 20 distinct dimensions, including logic reasoning, attribute recognition, and social relation understanding. 

\textbf{MME} provides a holistic evaluation across 14 subtasks categorized into perception and cognition. The perception track covers object counts, positions, and colors, while the cognition track tests commonsense reasoning, numerical calculation, and code translation. It is specifically designed to identify common pitfalls in Large Multimodal Models (LMMs), such as hallucination and lack of basic visual grounding.

\textbf{SEED-Bench} is a large-scale benchmark consisting of thousands of multiple-choice questions that cover both image and video understanding. It is organized into 12 evaluation dimensions, such as spatial awareness, temporal reasoning, and action recognition. The benchmark provides a unified framework to assess how well models generalize across different visual modalities and reasoning types.

\textbf{ScienceQA} is a multi-modal dataset comprising diverse science questions spanning three subjects (Natural Science, Language Science, and Social Science) and 26 topics. Most questions are accompanied by visual contexts and detailed explanations. It serves as a rigorous test for the model's high-level reasoning capabilities and its ability to synthesize knowledge from textbooks and diagrams.

\textbf{VQAv2} is the industry-standard benchmark for open-ended visual question answering on natural images. Building upon the original VQA dataset, it incorporates counter-examples to reduce language bias, requiring models to rely more heavily on visual evidence rather than statistical patterns in the text. It remains a primary metric for assessing basic visual-linguistic alignment.

\textbf{Q-Bench} focuses specifically on the ``low-level'' vision capabilities of multimodal models, such as assessing image quality, clarity, and aesthetic appeal. Unlike standard VQA, which focuses on high-level semantics (e.g., ``What is the cat doing?''), Q-Bench asks about distortions, noise, and lighting, evaluating whether the model perceives the technical attributes of an image correctly.

\subsubsection{Compared Methods}
We compare ETC with four representative visual-token compression baselines.

\textbf{MQT-LLaVA}~\cite{hu2024matryoshka} uses a query transformer with $M$ latent query tokens to compress visual features into a shorter token sequence. During training, it randomly samples a prefix of length $m \le M$ and keeps only the first $m$ query tokens while dropping the remaining ones. At inference time, the model directly uses the first $m$ compressed tokens under the target token budget, and in the reported setting it can be compressed to as few as 2 tokens.

\textbf{LLaMA-VID}~\cite{li2024llama} represents each visual input using two types of tokens: a context token and a content token. The context token is generated through cross-modal interaction between visual features and the user instruction, while the content token is obtained by pooling visual features to preserve compact visual content. The compressed visual representation is then formed by these two tokens and projected into the LLM input space, so its basic compressed form uses 2 tokens.

\textbf{VoCo}~\cite{ye2025voco} inserts a small number of Vision Compression (VoCo) tokens between visual tokens and text tokens in the LLM. It modifies the attention pattern so that VoCo tokens attend to visual tokens, while text tokens interact with visual information through the VoCo tokens. Training is performed with attention distillation, so that the activations on VoCo tokens learn to absorb information from the original visual-token sequence. In the extreme setting, VoCo can compress an image to a single token.

\textbf{QueCC}~\cite{li2025inference} performs query-dependent visual-token compression by injecting the user query into visual features before compression. It then combines spatial downsampling with cross-attention to map the original visual-token sequence into a very small number of compressed tokens, which are projected into the LLM input space. Its method is designed for extreme compression regimes and can reduce the visual input to as few as 1 token.

\subsubsection{Training parameters}
We implement the ETC framework across two representative architectures to validate scalability: LLaVA-1.5-7B \footnote{\url{https://huggingface.co/liuhaotian/llava-v1.5-7b}} and Qwen3-VL-2B \footnote{\url{https://huggingface.co/Qwen/Qwen3-VL-2B-Instruct}}. The training process is conducted on NVIDIA GeForce RTX 3090 GPUs using the DeepSpeed and MS-Swift frameworks. For both models, we freeze the vision tower while optimizing the cross-modal projector and the LLM backbone. Unless otherwise stated, the main experiments use a single learnable compressed token ($M=1$) and apply ETC supervision at the final layer of the LLM backbone. Consistent with the method formulation, we set the instruction-aware weighting strength to $\alpha = 0.6$ and the ETC sufficiency loss weight to $\lambda = 10^{-5}$.

The following table summarizes the hyperparameter configurations for both base models. While LLaVA-1.5-7B undergoes full fine-tuning with a global batch size of 128 to establish a robust baseline, Qwen3-VL-2B is adapted via LoRA for efficient scaling.

\subsubsection{Token-Budget Analysis}
\label{app:token_budget_analysis}
We further study how ETC behaves under different token budgets on SQA and TextVQA. As shown in Figure~\ref{fig:etc_token_scaling}, SQA reaches near-saturated performance with very few tokens: the score increases from 84.4 with 1 token to 85.7 with 8 tokens, which is already close to the full-model result of 85.77. Adding more tokens yields only minor gains. In contrast, TextVQA improves more steadily as the token budget increases, with accuracy continuing to rise from 1 to 8 to 32 to 64 tokens before largely flattening afterward (73.89 to 74.15). This difference suggests that tasks with different information complexity require different compression budgets. Overall, simpler tasks saturate early (around 8 tokens), while more challenging tasks benefit from larger token budgets (around 32--64), with clear diminishing returns in both cases.

\subsubsection{Visualization}
\begin{figure*}[htbp]
	\vskip 0.2in
	\centering
	\centerline{\includegraphics[scale=0.6]{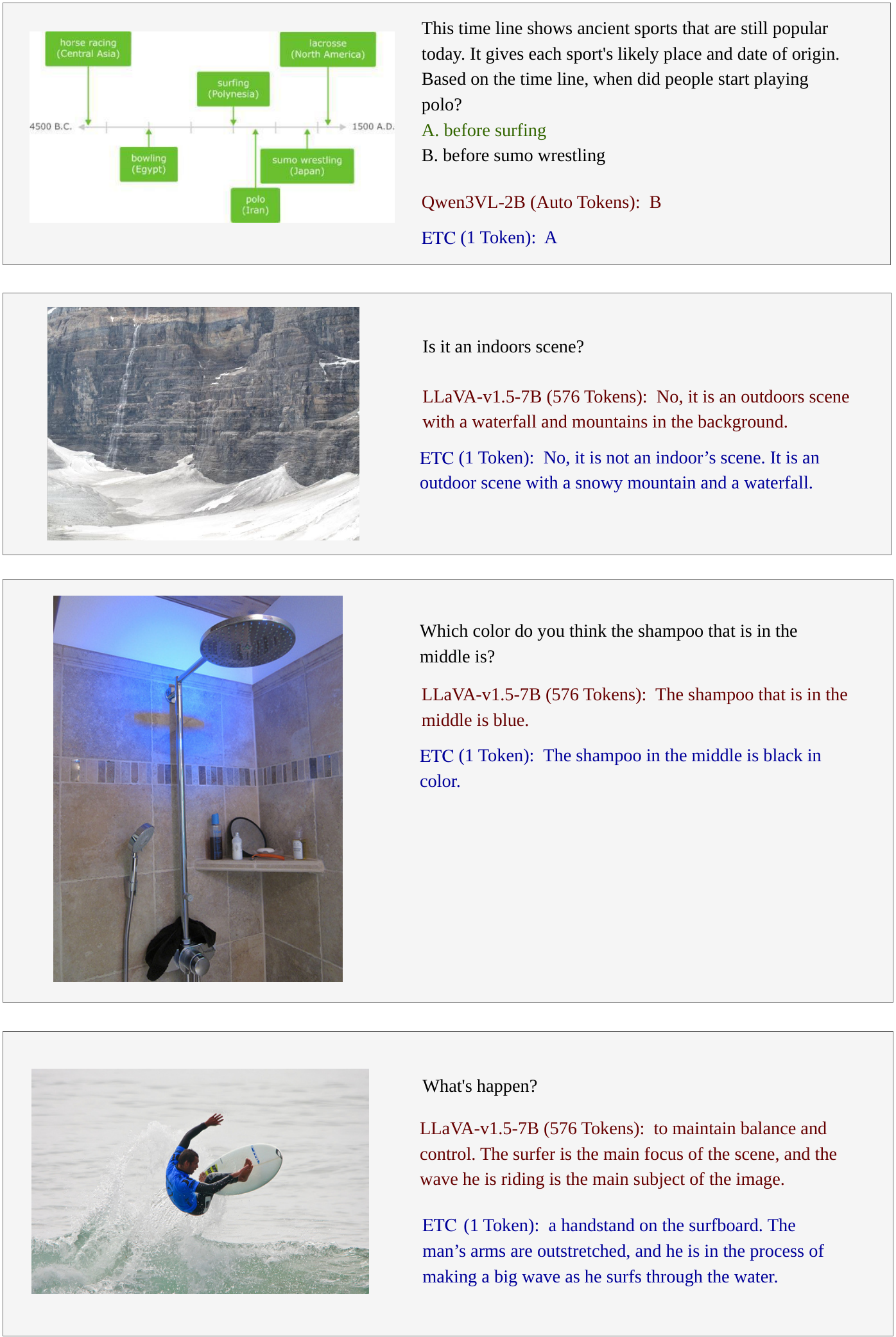}}
	\caption{
		Qualitative comparison of ETC with LLaVA-v1.5-7B and Qwen3VL-2B.
	}
	\label{fig:question}
\end{figure*}

We present qualitative evaluations across diverse visual reasoning tasks to further assess ETC's multimodal capabilities (Figure~\ref{fig:question}). Even under a 1-visual-token budget, ETC demonstrates comparable performance to token-heavy models in natural scene understanding (Question 2). Furthermore, ETC shows accurate instance-level recognition by identifying the liquid color (Question 3) and produces comprehensive, context-aware descriptions for extended tasks (Question 4). However, under a 1-visual-token budget, the information that can be preserved remains limited, which may lead to the loss of fine-grained details. For instance, in Question 1, ETC failed to capture these details and produced an incorrect answer. In summary, ETC is an efficient visual compression approach that retains key region-level information. Nevertheless, its capacity for fine-grained tasks can be further strengthened in future work.

\end{document}